\documentclass[runningheads]{llncs}

\usepackage{graphicx}
\usepackage[table]{xcolor}
\usepackage{collcell}
\usepackage{hhline}
\usepackage{pgf}
\usepackage{multirow}
\def\colorModel{hsb} 

\newcommand\ColCell[1]{
  \pgfmathparse{#1<12.5?1:0}  
    \ifnum\pgfmathresult=0\relax\color{white}\fi
  \pgfmathsetmacro\compA{0}      
  \pgfmathsetmacro\compB{#1/25} 
  \pgfmathsetmacro\compC{1}      
  \edef\x{\noexpand\centering\noexpand\cellcolor[\colorModel]{\compA,\compB,\compC}}\x #1
  } 
\newcolumntype{E}{>{\collectcell\ColCell}m{2cm}<{\endcollectcell}}  

\begin{document}
\title{Classification of Breast Cancer Histology using Deep Learning}
%
%
\author{Aditya Golatkar \and
Deepak Anand \and
Amit Sethi}
\authorrunning{A. Golatkar et al.}
%
\institute{Indian Institute of Technology Bombay, Mumbai, India
\email{\{chelsea.aditya,deepakanandece,amitsethi\}@gmail.com}}
\maketitle              
\begin{abstract}
Breast cancer is a major cause of death among women worldwide. Hematoxylin and eosin (H\&E) stained breast tissue samples from biopsies are observed under microscopes for primary diagnosis of breast cancer. In this paper, we propose a deep learning-based method for classification of H\&E stained breast tissue images released for BACH challenge~\cite{BACH} by fine-tuning Inception-v3 convolutional neural network (CNN)~\cite{inceptionv3}. These images are to be classified into four classes -- (i) normal tissue, (ii) benign lesion, (iii) \textit{in situ} carcinoma and (iv) invasive carcinoma. Our strategy is to extract patches based on nuclear density and rejecting patches that are not rich in nuclei, e.g. from non-epithelial regions. This allowed us to discard uninformative regions of the images as compared to random or grid sampling, because visual signs of tumors are most evident in the epithelium. Every patch with high nuclear density in an image is classified in one of the four above mentioned categories. The class of the entire image is determined using majority voting over the nuclear classes. We obtained an average accuracy of 85\% over the four classes and 93\% for non-cancer (i.e. normal or benign) vs. malignant (\textit{in situ} or invasive carcinoma), which significantly improves upon a previous benchmark~\cite{benchmark}.

\end{abstract}
\section{Introduction}
In 2012, breast cancer caused 522,000 deaths worldwide along with 1.68 million new cases~\cite{cancerStats}. Early diagnosis of the disease and proper treatment is essential to improve the survival rates. Examination of breast tissue biopsy using hematoxylin and eosin (H\&E) stain plays a crucial role in determining the type of lesion for primary diagnosis. Hematoxylin stains the nuclei purple and eosin stains the cytoplasm pinkish. This staining helps the pathologist identify the grade of carcinoma, which in turn determines the type of treatment to be provided to the patient. In this work we designed a deep learning based method to classify breast cancer slides. This work is part of our entry in the BACH challenge~\cite{BACH}. Solutions to this problem can potentially be used to reduce diagnoses errors, increase the throughput of pathologists, or be used in second opinion or teaching tools.

Tumors are believed to progress in phases. Normal breast tissues have large regions of cytoplasm (pinkish regions) with a dense cluster of nuclei forming glands in H\&E stained slides (Fig. 1-A). Benign lesion consists of multiple adjacent clusters of small-sized nuclei (Fig. 1-B). Unchecked benign lesions can progress to \textit{in situ} carcinoma in which the size of nuclei in the clusters increases and nucleoli within the nucleus become prominent, but the tumor seems to be circumscribed in round clusters while losing some of their glandular appearance (Fig. 1-C). In invasive carcinoma the enlarged nuclei break their clustered structure and spread to the nearby regions in fragments (Fig. 1-D). Carcinoma images have a high nuclear density with an absence of structure in inter-nuclear arrangement as compared to \textit{in situ} carcinoma images, which still have a preserved inter-nuclear structure.

The recent success of CNNs for natural image classification has inspired others and us to use them on medical images, e.g. for histopathology image classification. Spanhol \textit{et al}.~\cite{Spanhol} used an ImageNet~\cite{imagenet} based CNN architecture for classifying benign and malignant tumors. They extracted patches of sizes 32$\times$32 and 64$\times$64 from the images to train their CNN. In their results they showed that the accuracy of their CNN decreases with increase in magnification. Ciresan \textit{et al}.~\cite{ciresan}, who won the ICPR 2012 mitosis detection contest, trained a CNN on 101$\times$101 size patches extracted from images. This enabled them to analyze nuclei of different sizes. Cruz-Roa \textit{et al}.~\cite{roa} trained a CNN on 100$\times$100 size patches extracted out of whole slide images. They addressed the problem of detecting invasive carcinoma regions in whole slide images. Their CNN was able to extract structural as well as nucleus-based features. Their method established the-state-of-the-art by achieving an F1-score of 0.78. A recently proposed method by Ara\'{u}jo \textit{et al}~\cite{benchmark} addressed the problem of classifying H\&E stained images as normal, benign, \textit{in situ}, or invasive carcinoma. In their approach  they normalized the images using the method proposed in~\cite{Normalize}. They extracted 512$\times$512 patches from the normalized images to train their proposed CNN architecture. They also trained a CNN+SVM classifier for patch classification. Dataset augmentation was also performed by mirroring and rotating the patches. Images were classified by combining the patch probabilities using i) majority voting, ii) maximum probability and iii) sum of probabilities.
%
%
%
%

Challenges in the BACH dataset include vast areas in images without any epithelium (where the cancer starts) and areas of seemingly intermediate visual patterns between two neighboring classes.

\begin{figure}
\centering
\includegraphics[width=40mm,height=40mm]{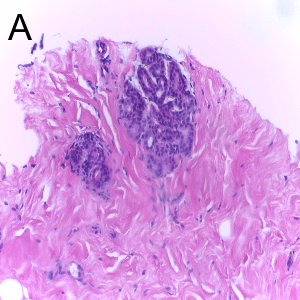}
\includegraphics[width=40mm,height=40mm]{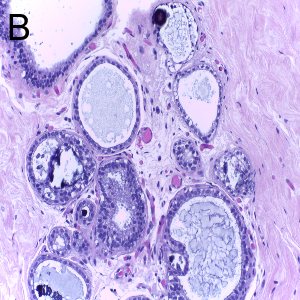}
\includegraphics[width=40mm,height=40mm]{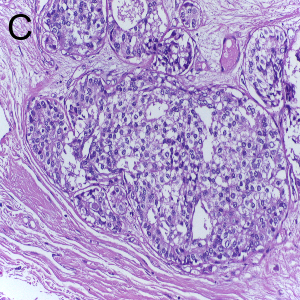}
\includegraphics[width=40mm,height=40mm]{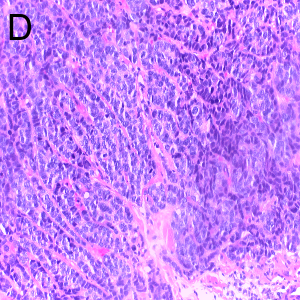}
\caption{Examples of H\&E stained images from the BACH challenge: A) normal tissue, B) benign lesion, C) \textit{in situ} carcinoma and D) invasive carcinoma. Hematoxylin stains the nuclei purple while eosin stains the stroma pink.}
\end{figure}

\section{Dataset}
%
The Breast Cancer Histology Challenge (BACH) 2018 dataset consists of high resolution H\&E stained breast histology microscopy images from ~\cite{BACH}. These images are RGB color images of size $2048\times 1536$ pixels. Each pixel covers $0.42 \mu m \times 0.42 \mu m$ of tissue area. The images in this dataset are annotated by two medical experts and cases of disagreement among the experts were discarded. The images are to be classified into four categories : i) normal tissue, ii) benign lesion, iii) \textit{in situ} carcinoma, and iv) invasive carcinoma as per the agreed upon diagnosis of the two experts. The dataset contains 100 images in each category amounting to a total of 400 images. For our experiments we have used 75 random images from each category for training and the remaining 25 images for validation. Thus in total, we used 300 images for training and 100 images for validating our method.

\section{Methods}
In this section we describe the details of our methods. First we will discuss about the novelty of our pre-processing method. Then we will discuss the details of patch-level classifier using transfer learning~\cite{transfer}. Finally we will explain the aggregation policy for generating the image-level classification, from patch-level classifier.

\subsection{Nuclei-based patch extraction}
CNNs trained on only a few hundred whole H\&E stained images of size 2048$\times$1536 pixels are prone to poor generalization due to overfitting. So, it is clear that CNNs have to be trained and applied on patches rather than whole images. This opens up the question of how the patches should be sampled from the whole slide images. The shape and size of a nucleus along with its surrounding structure is essential for accurate tissue classification. With this line of thinking in mind, it has previously been proposed to extracted patches centered at nuclei from the given H\&E stained images~\cite{SPIE17}. These extracted patches are also flipped horizontally and vertically, shifted horizontally and vertically and also rotated randomly within 180 degree for data augmentation. These methods not only help in increasing the dataset size by at least 8 folds but also makes our model more robust.

For patch extraction we divided the image into $299\times 299$ pixels patches with a 50\% overlap. We chose this patch size for two reasons. First, because fine-tuning Inception-v3 -- our base architecture -- requires images of that size, and we wanted to avoid inaccuracies that can come due to rescaling of images due to bilinear or cubic interpolations used in image resizing. Secondly, this patch size ensures that the CNN extracts nuclei based features along with features of inter-nuclear arrangements.

However, we do not use all the patches which are extracted from the image. Instead we only choose to keep those patches which have high nuclear density and discard patches that mostly cover stroma (with sparsely located nuclei) in majority of their area. To extract epithelial patches dense in nuclei we first compute a mask that identifies bluish pixels for each H\&E stained image by comparing  the ratio of blue and red channel intensities with to an appropriate threshold. By trial and error on the training images, we arrived at a threshold of 1.587. For each patch we define a blue density metric as the proportion of bluish pixels in the patch. All patches with more than $2\%$ bluish pixels were kept, and the rest were discarded. 
%
%
%
%
One challenge that we faced was that a few H\&E stained images in the given dataset have a large part of their area filled with stroma. In such cases very few pixels will be bluish and the image may not yield any patches for analysis. To overcome this problem, we first sort the extracted patches ($>2\%$ bluish pixels) in decreasing order of the proportion of bluish pixels.Then we define a blue density metric as the proportion of bluish pixels in the whole image. For images with metric more than $1\%$ we keep all the patches for scoring with CNN. For images with metric in range $0.5\%-1\%$,  $0.1\%-0.5\%$ and $<0.1\%$ we extract 10, 5 and 1 patch respectively for scoring with CNN. The reason for choosing less patches from such images is that stroma does not provide the model with any significant information regarding the type of tumor present in the image. Images with large regions of stroma are inconclusive and can even cause medical experts to have divided opinions. Fig 2 shows a benign image taken from the dataset along with its mask image. It can be seen that our model only extracts patches from regions dense with nuclei.
Each of the patch has the same class label as the original image.
%
%
\begin{figure}
\centering
\includegraphics[width=50mm,height=50mm]{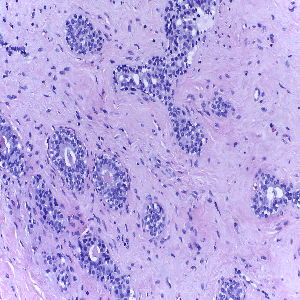}
\includegraphics[width=50mm,height=50mm]{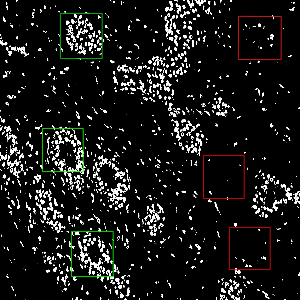}

\caption{A sample image with H\&E stained benign lesion (left), and its map of bluish pixels (right), with bounding boxes of accepted (green) and rejected patches (red).}
\end{figure}

\subsection{Transfer learning for patch-wise classification}
%
As mentioned earlier, the paucity of training images prevents us from training Inception-v3 from scratch with random initialization~\cite{inceptionv3}. Therefore, we employed transfer learning~\cite{transfer} and only fine-tuned Inception-v3 pre-trained on the ImageNet dataset~\cite{imagenet}. However, we have made some modifications to the Inception-v3 architecture. We removed the fully connected layer at the top of the network and added a global average pooling layer, followed by a fully connected layer with 1,024 neurons, and finally a softmax classifier with 4 neurons. Our training process had two stages. In the first stage we froze the convolutional layers and only trained the top (newly added) layers. In the second stage we fine-tuned the last two inception blocks along with the top layers. We used a Keras-based implementation of Inception-v3~\cite{inceptionv3}. We used RMSProp optimizer for 25 epochs for the first stage and SGD optimizer for the second stage with a learning rate of 0.0001 and momentum of 0.9 for 50 epochs. Disease label of an image was given to all its patches.  And, although this can lead to erroneous labels, because of the strategy to sample only nucleus-rich patches and the curation of the BACH dataset by its organizers, we expect that the patch-level labels are largely correct.



\subsection{Decision aggregation from patches to patients}
We combined the patch-based predictions using majority voting to determine the class of the entire image. In majority voting, the class of the entire image is the class to which maximum number of patches extracted from that image belong. In case of a tie we use the following precedence order to classify the image: i) invasive, ii) \textit{in situ}, iii) benign and iv) normal. The reason behind using this order was to avoid false negatives for a more dangerous disease class. 

\section{Results}
We employed two fold validation to evaluate our method by using patient-level accuracy on a held-out set of patients, with 25 such patients in each of the four classes. Accuracy for each class is defined as the ratio of correctly classified images to the total number images for that class in the validation set. Along with the four class classification we have also evaluated the accuracy of our method for identifying carcinoma images against non-carcinoma images. The non-carcinoma class consists of benign and normal images while the carcinoma class consists of \textit{in situ} and invasive carcinoma images. Along with the image-wise classification we have also computed the patch-wise accuracy for the bluish patches.
%
%
%
%
\subsection{Patch-level accuracy}
After performing the above-mentioned experiment we got an average patch-wise accuracy of 79\% across all the four classes. This metric compares favorably to a previous benchmark of 66.7\%~\cite{benchmark}. We attribute our success to nuclei-based patch extraction, which takes away lots of variability in patch appearance on which the CNN has to learn to be invariant. 

\subsection{Image-level accuracy}
We used majority voting to fuse the results obtained from the patch classification to predict the class of the entire image. The confusion matrices are shown in Tables 1 and 2.

\begin{table}
\centering
\caption{Four class confusion matrix}
\newcommand\items{4}   
\arrayrulecolor{white} 
\noindent\begin{tabular}{cc*{\items}{|E}|}
\multicolumn{1}{c}{} &\multicolumn{1}{c}{} &\multicolumn{\items}{c}{Actual} \\ \hhline{~*\items{|-}|}
\multicolumn{1}{c}{} & 
\multicolumn{1}{c}{} & 
\multicolumn{1}{c}{Normal} & 
\multicolumn{1}{c}{Benign} &
\multicolumn{1}{c}{\textit{In situ}} &
\multicolumn{1}{c}{Invasive} \\ \hhline{~*\items{|-}|}
\multirow{\items}{*}{\rotatebox{90}{Predicted}} 
&Normal  & 20   & 1  & 3 &  0 \\ \hhline{~*\items{|-}|}
&Benign  & 4   & 23  & 1  & 1 \\ \hhline{~*\items{|-}|}
&\textit{In situ}  & 1   & 1   & 20  & 2 \\ \hhline{~*\items{|-}|}
&Invasive  & 0   & 0   & 1  & 22 \\ \hhline{~*\items{|-}|}
\end{tabular}
\end{table}

\begin{table}
\centering
\caption{Two class confusion matrix}
\newcommand\items{2}   
\arrayrulecolor{white} 
\noindent\begin{tabular}{cc*{\items}{|E}|}
\multicolumn{1}{c}{} &\multicolumn{1}{c}{} &\multicolumn{\items}{c}{Actual} \\ \hhline{~*\items{|-}|}
\multicolumn{1}{c}{} & 
\multicolumn{1}{c}{} & 
\multicolumn{1}{c}{Non-Carcinoma} & 
\multicolumn{1}{c}{Carcinoma} \\ \hhline{~*\items{|-}|}
\multirow{\items}{*}{Predicted} 
&Non-Carcinoma  & 48   & 5   \\ \hhline{~*\items{|-}|}
&Carcinoma  & 2   & 45   \\ \hhline{~*\items{|-}|}
\end{tabular}
\end{table}

We can see in Table 1 that the our model confuses the normal class with benign. This can be attributed partly to the high similarity between benign and normal images in the dataset. Similarly, images of \textit{in situ} images carcinoma are confused with normal images due to a similar reason. The overall image-level accuracy was 85\%, which was higher than the patch-level accuracy due to the voting strategy.

Table 2 shows the image-level accuracy for non-carcinoma vs. carcinoma. The non-carcinoma super-class consisted of normal tissue as well as benign lesions. The carcinoma super-class was based on \textit{in situ} and invasive carcinomas. Accuracy on this task was 93\%.

A comparison of our proposed method with a previous benchmark is shown in in Table 3.
\begin{table}
\centering
\caption{Comparison of our results with a previous benchmark~\cite{benchmark} using same dataset.}\label{tab1}
\arrayrulecolor{black}
\begin{tabular}{|l|c|c|}
\hline
Method &  Proposed method & Ara{\'u}jo \textit{et al.}~\cite{benchmark}\\
\hline
Pre-processing &  Nuclei based patch extraction & Grid Sampling\\
\hline
Patch Classifier &  Modified Inception-v3 & Custom CNN\\
\hline
Patch-wise accuracy (4 class) & 79\% & 67\%\\
\hline
Image-wise accuracy (4 class) & 85\% & 78\%\\
\hline
Image-wise accuracy (2 class) & 93\% & 83\%\\
\hline
\end{tabular}
\end{table}

\section{Conclusion and Discussion}
We developed a two-stage approach for classification of the H\&E tissue images in four classes normal, benign, \textit{in situ}, and invasive. One of our key contributions was a pre-processing technique to consider only \textit{relevant} regions from tissue images for training and testing. We also used transfer learning for training the patch-level classifier with a large neural network architecture (Inception-v3) by using weights pre-trained on ImageNet dataset. We have achieved a classification accuracy of 79\% at patch-level. The image-level classification was done by majority-voting over patch-level classification results, resulting in an average accuracy of 85\% over the four classes and carcinoma versus non-carcinoma classification accuracy of 93\%. These figures were significantly higher than those achieved by a previous state-of-the-art~\cite{benchmark}

As far medical imaging is concerned, the application of deep learning has two major challenges namely \textit{the size of the image} and \textit{lack of region-of-interest annotations} for a large number of images. Several approaches can be used to further tackle these challenges apart from our static algorithm for patch selection. Multi-column CNNs that examine the tissue at multiple resolutions can be tried. Additionally, graph-based approaches to explicitly model inter-nuclear arrangements can also be tried. To account for staining variation in H\&E stained images following approach can be used. First color-normalize the images using~\cite{SPIE17}, extract nuclei based patches,  estimate the patient class of each patch, and then aggregate the estimated class decision for each tissue-image. The idea behind using nuclear-based patches as opposed to randomly or densely located patches is to give primacy to identifying nuclear structure and inter-nuclear arrangements based on our preliminary understanding of pathology in~\cite{SPIE17}. Additionally, multiple instance learning-based approaches that work with weakly supervised data can also be tried.

\section*{Acknowledgments}

Authors thank Nvidia Corporation for donation of GPUs used for this research.

\bibliographystyle{splncs04}

\end{document}